\documentclass[10pt, conference, compsocconf]{IEEEtran}
\IEEEoverridecommandlockouts
\usepackage{cite}
\usepackage{amsmath,amssymb,amsfonts}
\usepackage{algorithmic}
\usepackage{graphicx}
\usepackage{textcomp}
\usepackage{xcolor}
\usepackage{url}
\usepackage{paralist}
\def\BibTeX{{\rm B\kern-.05em{\sc i\kern-.025em b}\kern-.08em
    T\kern-.1667em\lower.7ex\hbox{E}\kern-.125emX}}
\begin{document}

\title{Analyzing Team Performance with Embeddings from Multiparty Dialogues 
\thanks{This material is based upon work supported by the Defense Advanced Research Projects Agency (DARPA) under Contract No. W911NF-20-1-0008.} 
}

\author{\IEEEauthorblockN{Ayesha Enayet}
\IEEEauthorblockA{\textit{Department of Computer Science} \\
\textit{University of Central Florida}\\
Orlando, USA \\
ayeshaenayet@knights.ucf.edu}
\and
\IEEEauthorblockN{Gita Sukthankar}
\IEEEauthorblockA{\textit{Department of Computer Science} \\
\textit{University of Central Florida}\\
Orlando, USA \\
gitars@eecs.ucf.edu}
}

\maketitle
\begin{abstract}
Good communication is indubitably the foundation of effective teamwork.  Over time teams develop their own communication styles and often exhibit entrainment, a conversational phenomena in which humans synchronize their linguistic choices.  This paper examines the problem of predicting team performance from embeddings learned from multiparty dialogues such that teams with similar conflict scores lie close to one another in vector space.  Embeddings were extracted from three types of features: 1) dialogue acts 2) sentiment polarity 3) syntactic entrainment.   Although all of these features can be used to effectively predict team performance, their utility varies by the teamwork phase.   We separate the dialogues of players playing a cooperative game into stages: 1) early (knowledge building) 2) middle (problem-solving) and 3) late (culmination). Unlike syntactic entrainment, both dialogue act and sentiment embeddings are effective for classifying team performance, even during the initial phase.  This finding has potential ramifications for the development of conversational agents that facilitate teaming.

% Our motivation is to identify some ways to predict team performance at the initial stages of the dialogue; this can help identify the need for assistance from a virtual or human agent. We study this problem by employing various NLP techniques. In this regard, we also encode entrainment into a vector to compare effect of syntax level feature with semantic level feature. We use doc2vec as our encoding algorithm.  We found that evolution of sentiments over the teamwork stages best represents the team's performance.Also, the semantic features showed more promising results even at the starting stage of communication then the syntactic feature. 
\end{abstract}


\section{Introduction}
The aim of our research is to create agents who can assist human teams by intervening when  teamwork goes awry.  To do this, it is important to be able to rapidly assess the status of team performance through ``thin-slicing'', making accurate classifications from short behavior samples; Jung suggests that developing this capability would remove the need for developing continuous team monitoring systems\cite{jung2016coupling}.  Ambady and Rosenthal demonstrate that many types of social interactions remain sufficiently stable that even a small sample is meaningful at predicting long term outcomes, the most famous application of this theory being thin-slicing marital interactions to predict divorce outcomes~\cite{ambady1,ambady2}.  Rather than developing specific measures for predicting future team conflict, we demonstrate that an embedding grouping teams with similar conflict levels can be learned directly from multiparty dialogue.  An advantage is that this approach avoids the necessity of collecting advance data on team members, such as personality traits or training records.

This paper compares the performance of three types of embeddings extracted from: 1) dialogue acts, 2) sentiment polarity, and 3) syntactic entrainment; these features were selected based on previous work on team communications and group problem-solving.
Dialogue acts capture the interactive pattern between speakers in multiparty communication\cite{goo2018abstractive}.  During dialogue act classification, utterances are grouped according to their communication purpose.
%example team communication categories from Walton and Krabbe's typology include information seeking, information providing, negotiation, and deliberation.
Sentiment polarity measures the attitude or emotion of the speaker during conversation; it can be used to detect disagreement. Entrainment is the natural tendency of the speakers to
adopt a similar style during a conversation, causing them to achieve linguistic alignment.  There are several types of entrainment including lexical choice~\cite{reitter}, style~\cite{danescu-style}, pronunciation~\cite{pardo}, and many others~\cite{traum}.   Reitter and Moore demonstrated that syntactic entrainment, based on alignment of lexical categories, can be used to predict success in task-oriented dialogues~\cite{reitter}. 

Good team communication exhibits all these characteristics: greater emphasis on problem solving than arguing, positive sentiment, and communication synchronization~\cite{yang2020}. Our research was conducted on the Teams corpus~\cite{litman2016teams} which consists of player dialogue during a cooperative game.  One advantage of studying a clearly defined, time-bounded team task is that the dialogues can be divided into teamwork phases: 1) early (knowledge building) 2) middle (problem solving) and 3) late (culmination).  For thin-slicing, we seek to predict the team performance from the initial teamwork stages.  The Teams corpus includes team conflict scores, which measure the amount of disagreement that occurred during gameplay.  Our hypotheses are:
\begin{compactitem}
\item \textbf{H1}: an embedding leveraging  dialogue acts will be useful for classifying team performance at all phases since it directly detects utterances related to conflict (eristic dialogues).
\item \textbf{H2}: sentiment analysis will consistently reveal team conflict and thus be a good predictor of performance.
\item \textbf{H3}: the entrainment embedding will be predictive when the entire dialogue is considered, but will be less useful at analyzing early phases before entrainment has been established.
\end{compactitem}
Embeddings are mechanisms for mapping high-dimensional spaces to low-dimensions while only retaining the most effective structural representations, making it possible to apply machine learning on large inputs by representing them in the form of sparse vector.  This paper presents our approach for extracting embeddings from multiparty dialogues that encode team conflict.  The next section describes the rich literature on analyzing team communication and multiparty dialogues.

%On the other hand, the progress in natural language processing has shown ways to represent pragmatics and semantics of dialogues in an automated way.  In this study, our objective is to encode DA \& sentiment patterns into a vector to study its effect on performance and introduce a simple yet effective way of predicting team social outcomes without extensive feature engineering.

%We focus on the simplified automation of the task by avoiding getting too much in the dialogue's contents. 

\section{Related Work}
Team communication, both spoken or written, is a critical element of collaborative tasks and can be studied in a variety of ways. Semantic analysis centers on the meaning of utterances, while pragmatics involves identifying speech acts\cite{bird1997survey}; both analytic approaches are important and often occur in parallel. In many studies of team communication, this analysis is arduously done through hand coding the utterances. 

Parsons et al. \cite{parsons2008analysis} contrast two different schemes to code utterances in team dialogues as part of their long term research goal of developing a virtual assistant for human teams. Their comparison  illustrates the benefits and problems of the Walton and Krabbe typology~\cite{waltonkrabbe}, which includes categories for information-seeking, inquiry, negotiation, persuasion, deliberation, and eristic, but does not consider the context in which the utterance occurs.  The McGrath theory of group behavior~\cite{mcgrath} focuses on modes of operation: inception, problem-solving, conflict resolution, and execution.  When applying the McGrath theory of group behavior, utterance classification is modified by conversational context. %Unfortunately there is no straightforward way to map the two schemes to one.

Sukthankar et al.\ also used an explicit team utterance coding scheme towards the problem of agent aiding of ad hoc, decentralized human teams to improve team performance on time-stressed group tasks~\cite{Sukthankar-Organization2009}.  Unlike teamwork studies, we do not specifically map individual utterances to team communication categories, but leverage dialogue act classification models to identify features that are indicative of team conflict. 
%Our study exploits the ability of machine learning to extract conversational context using semantic classification. 
Shibani et al.\cite{shibani2017assessing} discussed some of the practical challenges in designing an automated assessment system to provide students feedback on their teamwork competency: 1) dialogue pre-processing, 2) assessing teamwork chat text, and 3) classifying teamwork dimensions. They evaluated the performance of rule-based systems vs. supervised machine learning (SVM) at classifying coordination, mutual performance monitoring, team decision making, constructive conflict, team emotional support, and team commitment.  Even with dataset imbalance, the SVM model generally outperformed the hand coded rules. Our proposed method can also be used to assist human teams by proactively warning them of deficiencies during the early phases of team tasks, without the onerous data labeling requirements.
%Denescu et al. s  language signal exhibited by natural language during the conversation could provide valuable information about team dynamics\cite{danescu2012echoes}.

Other analytic techniques focus on linguistic coordination between speakers in groups.  For instance, Danescu et al.\ studied the effect of power differences on lexical category choices during goal-oriented discussion~\cite{danescu2012echoes}. This is one form of entrainment in which the speakers preferentially select function-word classes used by other group members.  Our paper uses a dataset (Teams corpus), that was created to study entrainment in teams~\cite{litman2016teams}. Rahimi and Litman demonstrated a method for learning an entrainment embedding to predict team performance \cite{rahimi2020entrainment2vec}; we use a modified version of their technique to express syntactic entrainment.  However since entrainment develops over time, we compare the performance of entrainment at early vs.\ late task phases. Furthermore, they only focused on syntactic/lexical features of utterances, not semantic.

%These team dialogues mostly occur in natural language, and studying these communications from the aspect of Natural language analysis yields informative results.  Dialogue Acts and sentiment polarities gives useful Semantic and pragmatic information. Encoding this information into spares numerical vectors can help in studying team dynamics

Sentiment analysis has been applied to the study of group dynamics; for instance, researchers have leveraged sentiment features to detect communities in social networks \cite{sawhney2017community,xu2011sentiment}. Our work demonstrates the utility of sentiment features towards predicting team conflict and show that the sentiment-based embedding is useful during all teamwork phases. We rely exclusively on the multiparty team dialogues; however there have been many attempts to predict team performance using other types of multimodal features.  TCdata, a team cooperation dataset, includes both audio and video recordings of teams performing cooperative tasks~\cite{liu2017analysis}.
Liu et al. explicitly extracted 159 features from team speaking cues, individual speaking time statistics, and face-to-face interaction cues to predict team performance on this dataset.

Several studies~\cite{yang2014prediction,omar2011developing} have shown team member personality traits to be useful predictors of conflict and team performance.
Yang et al. used individual personality traits to predict the performance of final year student project teams using neural networks~\cite{yang2014prediction}. Omar et al. developed a student performance prediction model that included both personality types and team personality diversity~\cite{omar2011developing}.  Even though these additional data sources can be highly predictive, they are rarely available in real-world team scenarios, unlike multi-party dialogue which is often self-archived to preserve organizational memory.

\section{Method}
This section describes our procedure for computing embeddings using doc2vec~\cite{le2014distributed}, an unsupervised method that is used to create a vector representation of the team dialogue. We compare the performance of different possible inputs to doc2vec: 1) dialogue acts, 2) sentiment analysis, and 3) syntactic entrainment. 

\subsection{Dialogue Acts}
\begin{figure}
  \centering
  \includegraphics[width=\linewidth]{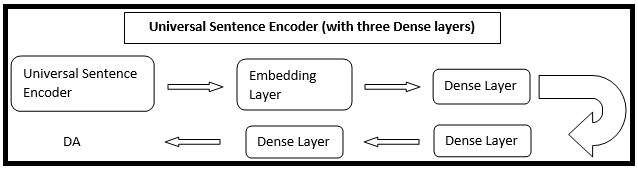}
  \caption{Dialogue Act Classifier Architecture.}
  \label{DA}
\end{figure}

Dialogue acts can be created from the semantic classification of dialogue at the utterance level to identify the intent of  the speaker. A transfer learning approach was used to tag utterances of the Teams corpus using the DAMSL (Discourse Annotation and Markup System of Labeling) tagset.  Figure~\ref{DA} shows the architecture of our dialogue act classifier, which was constructed using the Universal Sentence Encoder; we selected USE for its ability to achieve consistently good performance across multiple NLP tasks~\cite{cer2018universal}.
 There are two different variants of the model: 1) a transformer architecture, which exhibits high accuracy at the cost of increased resource consumption and 2) a deep averaging network that requires few resources and makes small compromises for efficiency. The former uses attention-based, context-aware encoding subgraphs of the transfer architecture. The model outputs a 512-dimensional vector. The deep averaging network works by averaging words and bigram embeddings to use as an input to a deep neural network. The models are trained on web news, Wikipedia, web question-answer pages, discussion forums, and the Stanford Natural Language Inference (SNLI) corpus, and are freely available on TF Hub.
 We selected the USE Transformer-based Architecture model with three dense layers and a softmax activation function. Figure \ref{DA} shows the architecture of our DA classification model, which achieves a validation accuracy of 70\%.
 
 The model was fine-tuned using the  Switchboard Dialogue Act Corpus (SwDA) dataset. SwDA is one of the most popular public datasets for DA classification. It consists of 1155 human-to-human telephone speech conversations, tagged using 42 tags from the DAMSL tagset.  Table \ref{tab:desdataset} shows the statistics of both SwDA and the Teams corpus.  

\begin{table}
\centering
\caption{Dataset Statistics}\label{tab:desdataset}
\begin{tabular}{|l|l|l|}
\hline
Dataset & \#Utterances& \#Tokens\\
\hline
SwDA & 200,052&19K\\
Teams Corpus & 110,206& 573,200\\
\hline
\end{tabular}
\end{table}
\begin{table*}
\centering
\caption{SwDA Dataset Sample}\label{tab:SwDAUtterences}
\begin{tabular}{|p{1cm}|p{6cm}|p{1cm}|p{4cm}|}
\hline
Speaker&Utterance&DA&Description\\
\hline
A&I don't, I don't have any kids. & sd& Statement-non-Opinion\\
     A&I, uh, my sister has a, she just had a baby, &sd& Statement-non-Opinion \\
     A& he's about five months old&sd& Statement-non-Opinion \\
     A&and she was worrying about going back to work and what she was going to do with him and -- &sd& Statement-non-Opinion \\
     A&Uh-huh. & b& Acknowledge\\
     A&do you have kids? &qy&Yes-No-Question \\
     B&I have three. &na&Affirmative non-yes Answer \\
     A&Oh, really? &bh&Backchannel in question form \\
\hline
\end{tabular}
\end{table*}
\begin{table*}
\centering
\caption{Teams Corpus Example}\label{tab:GitHubUtterences}
\begin{tabular}{|p{1cm}|p{6cm}|p{1cm}|p{4cm}|}
\hline
Speaker&Utterance&DA& Description\\
\hline
     A&Ok I'm going to&sd&Statement-non-Opinion \\
     A&shore up these two.&sd&Statement-non-Opinion \\
     B&Good move.&ba&Appreciation \\
     A&Then we got one and then I guess I can also& sd&Statement-non-Opinion\\
     A&Can I use my powers twice in one play&sd&Statement-non-Opinion \\
     C&Mm&b&Acknowledge (Backchannel) \\
     B&yes &ny&Yes answer \\
\hline
\end{tabular}
\end{table*}
Table~\ref{tab:SwDAUtterences} shows examples from the SwDA training dataset, and  Table~\ref{tab:GitHubUtterences} shows examples from Teams corpus. Each team dialogue generates a unique sequence where each element of the sequence represents the dialogue act of the corresponding utterance.  This sequence of dialogue acts is then used as an input to doc2vec algorithm to create the embedding.   

\subsection{Sentiment Analysis}
Another option is to represent the team dialogue as a series of changes in the emotional state of the team.  This can be done by applying sentiment analysis to the individual utterances.
Sentiment analysis is the task of predicting the emotion or attitude of the speaker; we are using the TextBlob python implementation~\cite{textblob} to determine sentiment polarity of each utterance in the dialogue. The polarities are float values which lies between -1 and 1 representing negative, positive and neutral sentiment. For each team the unique sequence of these polarities is used as input to doc2vec, where each element of the sequence represents the polarity of the corresponding utterance. This representation encodes transitions in the emotional state of the team across the duration of the task.

\subsection{Entrainment}\label{entrainment}
Entrainment is one form of linguistic coordination in which team members adopt  similar speaking styles during  conversation.  Here we evaluate the performance of a syntactic entrainment embedding based on Rahmi and Litman's \cite{rahimi2020entrainment2vec}'s work that encodes the propensity of subsequent speakers to make similar lexical choices.  Eight lexical categories were used: noun (NN), adjective (JJ), verb (VB), adverb (RB), coordinating conjunction (CC), cardinal digit (CD), preposition/subordinating conjunction (IN), and personal pronoun (PRP) . To calculate the entrainment between two speakers we follow the method proposed by Danescu et al. \cite{danescu2012echoes} shown in Equation \ref{eq1}. $Ent_c(x,y)$ is the entrainment of speaker $y$ to speaker $x$, $c$ is the lexical category, $e_{yx^c}$ represents the event where speaker $y$ utterance immediately follows the  speaker $x$ utterance and contains $c$, ${e_x^c}$ is the event when utterance (spoken to y) of speaker $x$ contains $c$. 
\begin{equation}
    \label{eq1}
  Ent_c(x,y)=p(\frac{e_{yx^c}}{e_x^c})-p(e_{yx^c})
\end{equation}

The NLTK part-of-speech (POS) tagger was used to tag all the utterances with their respective lexical categories.  A directed weighted graph was generated for each dialogue linking speakers with positive entrainment. The structure of this graph encodes the entrainment relationships between team members.  To translate the graph into a feature representation, six graph centrality kernel functions were applied to represent each node of the team graph. The kernel functions are: (1) PageRank (2) betweenness centrality (3) closeness centrality (4) degree centrality (5) in degree centrality (6) Katz centrality.  To create the final team representation, the vectors of individual nodes were averaged, and doc2vec was applied to create the embedding. This method corresponds to the Kernel version of Entrainment2Vec~\cite{rahimi2020entrainment2vec} and achieves comparable performance when applied to the whole dialogue.

%Table \ref{tab:KernalAlgo} gives the brief overview of all the kernel functions. To generate a team level feature vector, we average vectors of the individual nodes.

Our implementation is slightly different from that of \cite{rahimi2020entrainment2vec} and \cite{danescu2012echoes} in two aspects. First, we are using the NLTK POS tagger to assign lexical categories to the utterances instead of using LIWC-derived categories. Second, we are using six graph kernel algorithms instead of ten. We observed that using more graph kernel functions on graphs that  consist of three to four team members does not improve performance.  The POS tagging reflects the sentence's syntactic structure; we have carefully selected the POS categories that are consistent with the conventional English part of speech categories used by \cite{rahimi2020entrainment2vec} and \cite{danescu2012echoes}. While calculating the entrainment, we do not consider the actual word and its context; therefore, this embedding only captures syntactic features, not semantics. 
%\begin{table*}
%\centering
%\caption{Description of Kernel Functions}\label{tab:KernalAlgo}
%\begin{tabular}{|p{3cm}|p{6cm}|p{4cm}|}
%\hline
%Kernel Function&Definition&Description\\
%\hline
%     Pagerank&Pagerank algorithm ranks the node based on the quality and number of the links towards it. &The basic intuition is, the more the number of links the node has, the more important it is.  \\
%     Betweenness centrality&It measures the centrality of the node based on the shortest path. &The high betweenness centrality score of the node reflects that more information passes through it. \\
%     Closeness centrality&It measures a node's centrality as the reciprocal of the sum of the length of the shortest paths between the node with all the other nodes in the graph.  &The higher the closeness centrality score of the node, the closer it is with other nodes. The nodes with high closeness centrality score spread information efficiently. \\
%     Degree centrality&It is the measure of the number of connections the node has. It counts both incoming and outgoing connections.& It reflects the entrainment to or from the node.\\
%     In-degree centrality&It measures the number of the incoming connection to the node.&The score determines the number of influencers of the node.  \\
%     Katz centrality &It measures the relative degree of influence of the node based on the number of walks between the two nodes.   &It reflects the node's relative influence on its neighbor(s) and neighbor(s) of the neighbor.\\
%\hline
%\end{tabular}
%\end{table*}

\subsection{Doc2vec}
 Le and Mikolov\cite{le2014distributed} introduced doc2vec as an unsupervised learning algorithm to generate distributed vector representations of text of arbitrary size; it is inspired by the word2vec model\cite{mikolov2013distributed}.  They proposed two different models for learning numerical representations of text: 1) Distributed Memory Model of Paragraph Vectors (PV-DM) 2) paragraph vector with a distributed bag of words (PV-DBOW).

\textbf{Distributed Memory Model of Paragraph Vectors (PV-DM)} uses both word vectors and paragraph vectors to predict the next word. It attempts to learn paragraph vectors that can predict the word given different contexts sampled from the text.  The context size is a tuneable parameter, and a sliding window of arbitrary context size generates multiple context samples. Doc2vec works by averaging these word vectors and paragraph vectors to predict the next word. It employs stochastic gradient descent to learn word and paragraph vectors. The resultant paragraph vectors serve as a feature vector of the corresponding paragraph and can be used as an input to machine learning models like SVM and logistic regression.

\textbf{Paragraph vector with a distributed bag of words (PV-DBOW)} ignores the context words and attempts to predict randomly selected words from the paragraph. At each iteration of stochastic gradient descent, it classifies a randomly selected word from the sampled text window using paragraph vectors.

Instead of using doc2vec on the raw team dialogues, doc2vec was applied to the output of the dialogue act classifier, sentiment analysis, and syntactic entrainment.  This procedure enables us to disentangle the contribution of different elements of team communication at predicting conflict. 

\section{Dataset}
Our evaluation was conducted on the Teams corpus dataset collected by Litman et al.~\cite{litman2016teams}.  It contains 124 team dialogues from 62 different teams, playing two different collaborative board games.  The length of the dialogues varies from 291 to 2124 utterances. In addition to collecting dialogue data, the researchers administered surveys of team level social outcomes.
Team social outcome scores include task conflict, relation conflict, and process conflict scores. All these scores are highly correlated, and we are using process conflict z-scores to represent team performance. Jehn et al. have identified that low process conflict scores indicate good team performance and vice versa \cite{jehn2001dynamic}.
To study the problem of early prediction of team conflict, we divide each dialogue into three equal sections that correspond to the knowledge-building, problem solving, and culmination teamwork phases. Our final classification dataset consists of 12 patterns per dialogue, which are generated from applying the three methods (semantic, sentiment, syntactic) to the whole time period, as well as the initial, middle and final segments. 

Teams were divided into high performing and low performing teams based on their process conflict z-scores, and classification accuracy was measured.
 Doc2vec was used to generate the vector representation of all the patterns. Doc2vec comes in two different flavors: 1) Distributed Memory Model of Paragraph Vectors (PV-DM) and 2) Distributed Bag of Words version of Paragraph Vector (PV-DBOW). Through extensive experiments, we  identified that PV-DM with epoch size of 5, negative sampling 5, and window size 10 works best for our setting. By default, we only report results for PV-DM. Table \ref{tab:DMvsDBOW} shows the comparison of PV-DM \& PV-DBOW when applied to the complete dialogue. 
\begin{table}
  \caption{Doc2Vec Comparison}
  \centering
  \label{tab:DMvsDBOW}
  \begin{tabular}{|p{2cm}|p{3cm}|p{2cm}|}
    \hline
    &PV-DBOW&PV-DM\\
    \hline
    Dialogue Act & 57.89& 68.42\\
    Sentiment & 55.26& 78.94\\
    Entrainment & 55.26& 60.52\\
  \hline
\end{tabular}
\end{table}
We evaluated the performance of both logistic regression and the support vector machine (SVM) classifier on the full dialogue (shown in Table~\ref{tab:acccomp}); for the other experiments, the better performer, SVM, was used.

\begin{table}
  \caption{Comparison of Supervised Classifiers}
  \centering
  \label{tab:acccomp}
  \begin{tabular}{|p{2cm}|p{3cm}|p{2cm}|}
    \hline
    &Logistic Regression&SVM\\
    \hline
    Dialogue Act & 63.15& 68.42\\
    Sentiment & 71.05 & 78.94\\
    Entrainment & 63.15& 60.52\\
  \hline
\end{tabular}
\end{table}
\section{Results}
\begin{figure*}
  \centering
  \includegraphics[width=0.3\linewidth]{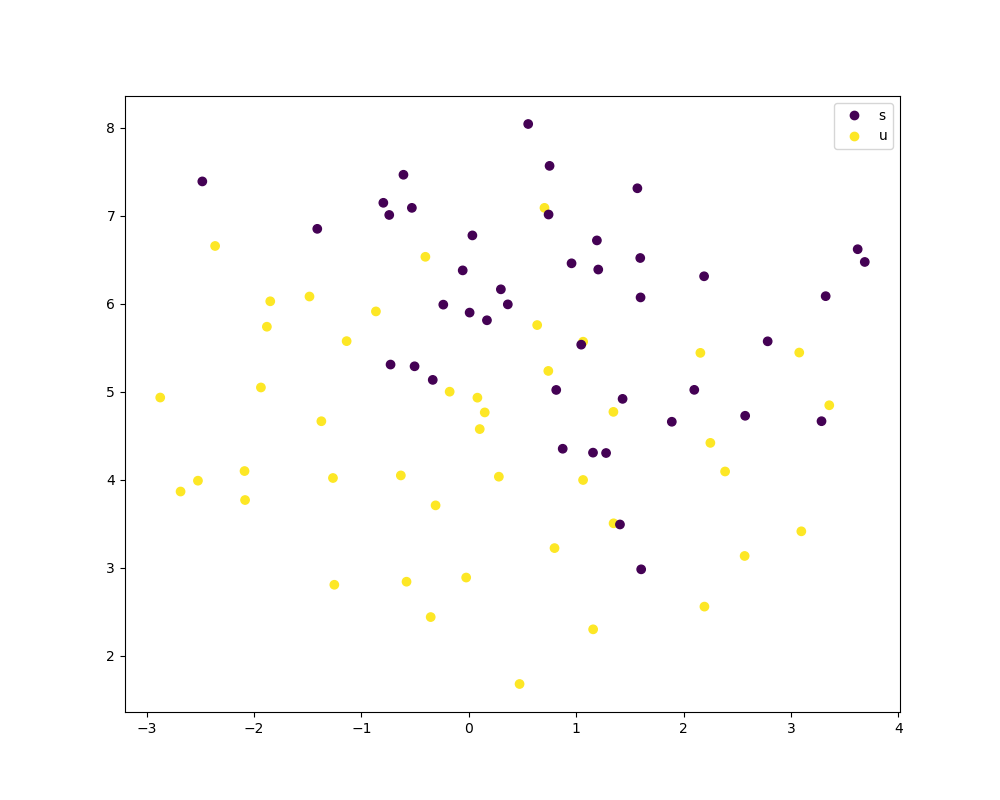}
  \includegraphics[width=0.3\linewidth]{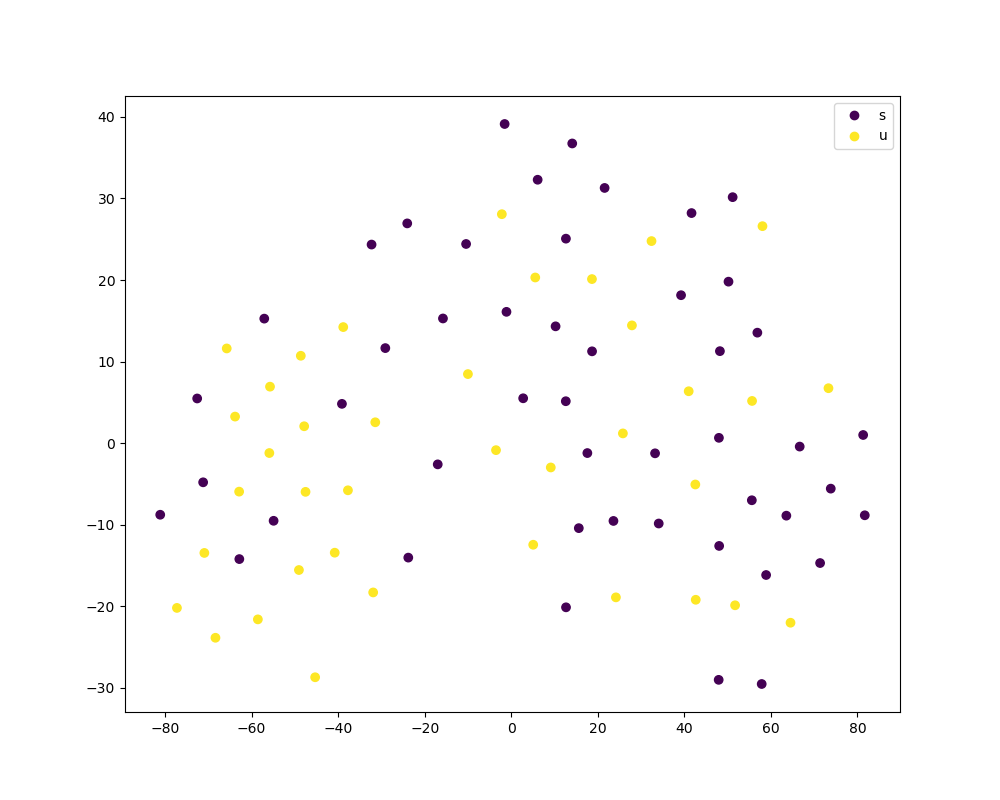}
  \includegraphics[width=0.3\linewidth]{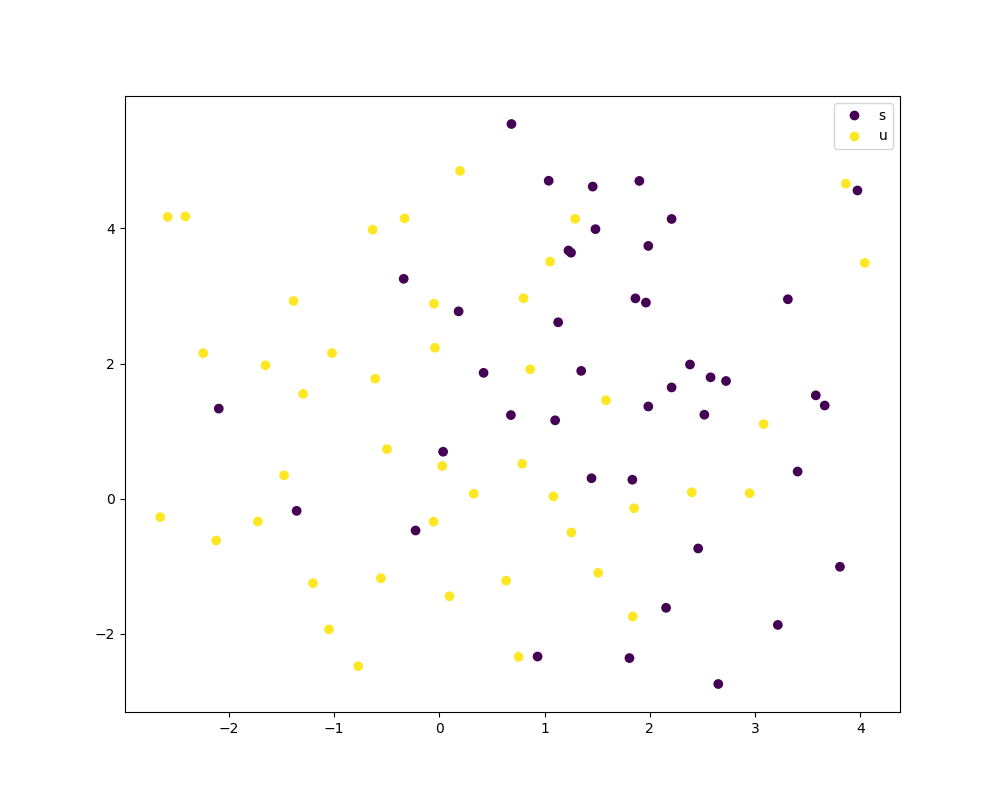}
  \caption{t-SNE representation of vectors in 2D, where 'S' represents the teams with low process conflict scores and 'U' represents the teams with high process conflict scores. Both sentiment (left) and dialogue act embedding (right) show a better class separation than entrainment (center).  Note that the axes have no explicit meaning.}
  \label{fig1}
\end{figure*}

Table \ref{tab:acc} presents the classification accuracy of the three embeddings on the whole dialogue. SVM exhibits the best classification accuracy of 78.94\% on sentiment based vectors, followed by dialogue act based vectors. Figure \ref{fig1} visually illustrates the effects of different embeddings.  By plotting the vectors in 2d  using t-Distributed Stochastic Neighbor Embedding (TSNE), we can observe the formation of two clusters, representing teams with high social outcomes and low social outcomes in the dialogue act and sentiment vectors, whereas the entrainment ones are intermixed.

\begin{table}
  \caption{Accuracy by Team Phase}
  \centering
  \label{tab:acc}
  \begin{tabular}{|p{1.5cm}|p{1.5cm}|p{1.5cm}|p{1.5cm}|}
    \hline
    Phase & DA &Sentiment &Entrainmemt\\
    \hline
    Whole & 68.42 & 78.94& 60.52\\
    Initial & 71.05 & 65.78& 42.10\\
    Middle & 73.68 & 65.78& 47.36\\
    End & 68.42 & 71.05 & 60.52\\
  \hline
\end{tabular}
\end{table}
Table~\ref{tab:acc} shows the accuracy of the conflict classifier across the duration of the games.  The sentiment classifier achieved the best accuracy when the whole dialogue was used and exhibited consistent performance across all team phases.  The dialogue act embedding was the best at the initial phase, making it a good choice for the ``thin-slice'' problem of rapidly diagnosing teamwork health from a small sample of utterances.  Syntactic entrainment lagged behind the sentiment and semantic analysis, but performance improved during the final phase.

\begin{figure*}
  \centering
  \includegraphics[width=0.3\linewidth]{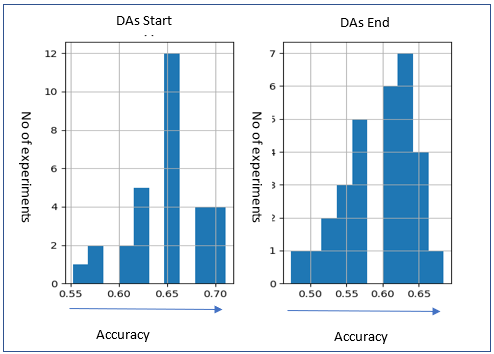}
  \includegraphics[width=0.3\linewidth] {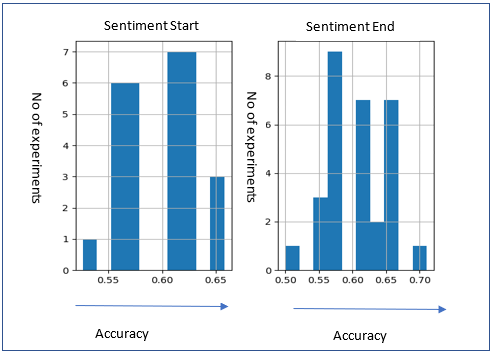}
  \includegraphics[width=0.3\linewidth]
  {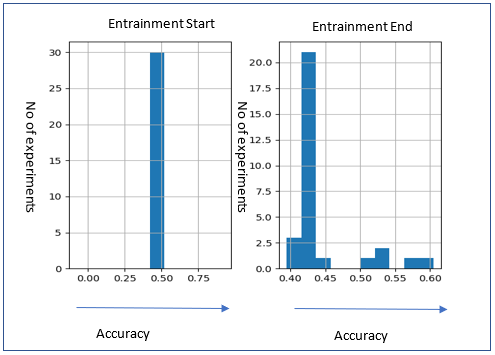}
  \caption{Distribution of embedding results for initial and final teamwork phases for dialogue acts (left), sentiment (middle) and entrainment (right)} 
  \label{fig4}
\end{figure*}

For statistical testing, we generated 30 results for each phase using each embedding.  Since some of the result distributions (Figure~\ref{fig4}) failed the D'Agostino-Pearson normality test, the Kolmogorov-Smirnov test was used for significance testing. The performance differences between each pair of embeddings were statistically significant ($p<0.01$).  However the differences between the  initial and end phase results for the sentiment and entrainment embeddings were not significant (Table~\ref{tab:sig2}).  
Semantic and sentiment based vectors outperformed the syntactic entrainment vectors at the classification task across all phases. 

%We observe that during the initial phase the entrainment vector showed very little variation in accuracy, while ase it achieved maximum accuracy of 60\%. 

\begin{table*}
  \caption{ Comparison of performance of all the three approaches at Knowledge Discovery \& Culmination Phase} 
  \centering
  \label{tab:sig2}
  \begin{tabular}{|cccccl|}
    \hline
    &\multicolumn{2}{c}{Knowledge Discovery}&\multicolumn{2}{c}{Culmination}&\\
    \hline
    &min & max&min&max&p-value\\
    Dialogue Act&0.552632&0.710526&0.473684&0.684211&\textbf{2.48e-05}\\
    Sentiment&0.526316&0.657895&0.500000&0.710526&0.455695\\
    Entrainment&0.4210&0.4210&0.394737&0.605263&0.594071\\
  \hline
\end{tabular}
\end{table*}
\section{Conclusion}
This study presents an evaluation of different embeddings for predicting team conflict from multiparty dialogue. Embeddings were extracted from three types of features:
1) dialogue acts 2) sentiment polarity 3) syntactic entrainment.  Results confirm the effectiveness of both sentiment (\textbf{H2}) and dialogue acts (\textbf{H1}).  However, experiments failed to confirm that classification based on syntactic entrainment signficantly improves over time (\textbf{H3}).  Although there are many other ways to measure linguistic synchronizaton, it seems less promising for integration into an agent assistance system.  The dialogue act embedding is strong during the initial phase making it a good candidate for diagnosing the health of team formation activity.  A continuous team monitoring agent assistant system might do better with sentiment analysis. 

In future work we plan to explore embeddings based on macrocognitive teamwork states, such as those in
the Macrocognition in Teams Model (MITM)~\cite{fiore2010}.  Drawing from research on externalized cognition, team cognition, group communication and problem solving, and collaborative learning and adaptation, MITM provides a coherent theoretically based conceptualization for understanding complex team processes and how these emerge and change over time.
It captures the parallel and iterative processes engaged by teams as they synthesize these components in service of team cognitive processes such as problem solving, decision making and planning. 

\section{Acknowledgement}
This material is based upon work supported by the Defense Advanced Research Projects Agency (DARPA) under Contract No. W911NF-20-1-0008. Any opinions, findings and conclusions or recommendations expressed in this material are those of the authors and do not necessarily reflect the views of DARPA or the University of Central Florida.

\bibliographystyle{IEEEtran}
\bibliography{IEEEfull}

\end{document}